\documentclass[a4paper,conference]{IEEEtran}

\usepackage{graphicx}
\usepackage{amsmath}
\usepackage{amssymb}
\usepackage{threeparttable}
\usepackage{array}
\usepackage{cite}

\newcommand{\remotenet}{ReMotENet}

\newcolumntype{P}[1]{>{\centering\arraybackslash}p{#1}}
\newcommand{\etal}{et al.}
\newcommand{\eg}{\textit{e.g.}}
\begin{document}
%
\title{TempNet: Temporal Attention Towards the Detection of Animal Behaviour in Videos}



%
\author{\IEEEauthorblockN{Declan McIntosh\IEEEauthorrefmark{1},
Tunai Porto Marques\IEEEauthorrefmark{1},
Alexandra Branzan Albu\IEEEauthorrefmark{1},
Rodney Rountree\IEEEauthorrefmark{2} and
Fabio De Leo\IEEEauthorrefmark{3}}
\IEEEauthorblockA{\IEEEauthorrefmark{1}Department of Electrical and Computer Engineering, \\
University of Victoria,
Victoria, British Columbia\\ Email: aalbu@uvic.ca}
\IEEEauthorblockA{\IEEEauthorrefmark{2}Department of Biology, University of Victoria, Victoria, British Columbia }
\IEEEauthorblockA{\IEEEauthorrefmark{3}Ocean Networks Canada, Victoria, British Columbia }}


\maketitle

\begin{abstract}
  Recent advancements in cabled ocean observatories have increased the quality and prevalence of underwater videos; this data enables the extraction of high-level biologically relevant information such as species' behaviours. Despite this increase in capability, most modern methods for the automatic interpretation of underwater videos focus only on the detection and counting organisms. We propose an efficient computer vision- and deep learning-based method for the detection of biological behaviours in videos. TempNet uses an encoder bridge and residual blocks to maintain model performance with a two-staged, spatial, then temporal, encoder. TempNet also presents temporal attention during spatial encoding as well as Wavelet Down-Sampling pre-processing to improve model accuracy. Although our system is designed for applications to diverse fish behaviours (i.e, is generic), we demonstrate its application to the detection of sablefish (\textit{Anoplopoma fimbria}) startle events. We compare the proposed approach with a state-of-the-art end-to-end video detection method (ReMotENet) and a hybrid method previously offered exclusively for the detection of sablefish's startle events in videos from an existing dataset. Results show that our novel method comfortably outperforms the comparison baselines in multiple metrics, reaching a per-clip accuracy and precision of 80\% and 0.81, respectively. This represents a relative improvement of 31\% in accuracy and 27\% in precision over the compared methods using this dataset. Our computational pipeline is also highly efficient, as it can process each 4-second video clip in only 38ms. Furthermore, since it does not employ features specific to sablefish startle events, our system can be easily extended to other behaviours in future works. 
\end{abstract}


%
\IEEEpeerreviewmaketitle

\section{Introduction}

Underwater optical systems in ecological and fisheries monitoring have become increasingly prevalent over the last six decades \cite{mallet2014underwater}. However, biologically-relevant advancements in automated video processing often do not mirror the advancements in video acquisition. Cabled ocean observatories have recorded and stored an excess of thousands of hours of underwater video \cite{heesemann2014ocean} to assess organism presence, abundance, and activity patterns. However, manually interpreting these data requires enormous amounts of time, representing tedious tasks that can lead to errors and inter-expert disagreements. Hence, there exists a need to develop semi- and fully-automated frameworks capable of enhancing \cite{porto2020l2uwe, mdpipaper, ancuti2012enhancing, akkaynak2019sea} and annotating \cite{toh2009automated,rauf2019visual, albuquerque2019automatic, zhang2020automatic, spampinato2008detecting} ever-expanding marine imagery datasets. \par

While significant efforts have been made to develop tools for the autonomous interpretation of underwater imagery in the last decade~\cite{aguzzi2020potential}, most works concentrate on methods that are limited to the identification and counting of specimens. However, complex ecological analyses typically require a more complex interpretation of such data. 
Video data streams provide the unique opportunity to obtain critical information on individual organism behaviour events, as well as their inter-specimen interactions. 
\begin{figure*}
\begin{center}
\includegraphics[width=1.0\linewidth]{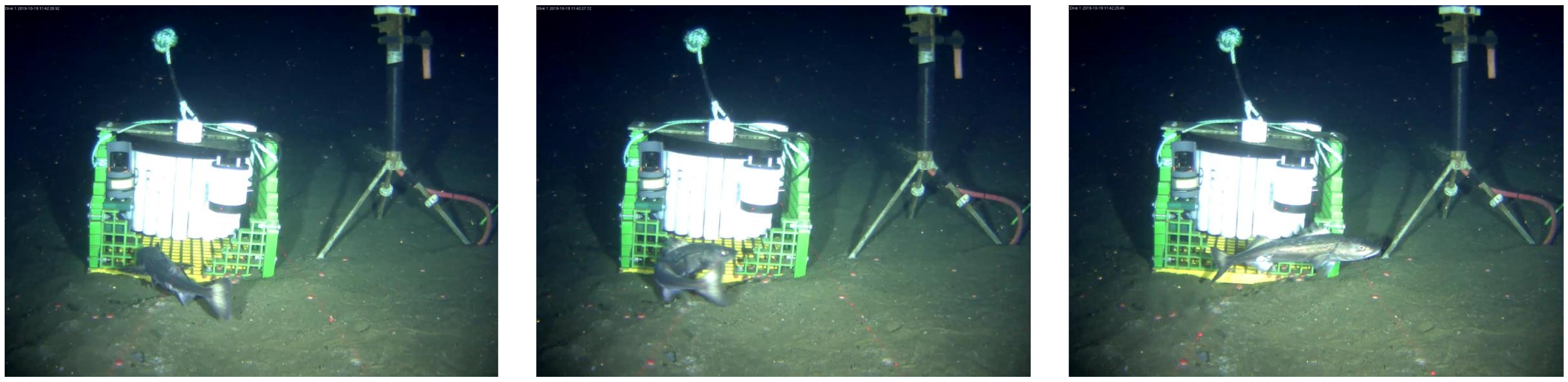}
\end{center}
   \caption{Three example frames from one clip containing a sablefish startle event occurring from interaction with man-made instruments. Frames were selected from before the beginning of the event, the middle of the event and after the event has concluded. The duration of this event at 5Hz is 3 frames.}
\label{fig:1-example}
\end{figure*}

An efficient method for the automatic detection of some of these behaviours  holds an enormous biological value, as it allows for a specific interpretation of long-term video recordings from offshore observatories \cite{rountree2019towards,heesemann2014ocean}, \textit{e.g.}, identifying predation events of a given species, rather than just its presence. As a result, this capability represents a change in the level of abstraction of machine learning- and computer vision-based interpretation of underwater data \cite{mcintosh2020movement}: from a narrow and context-agnostic counting of specimens to a broader and biologically complex identification of specific behaviours.\par

In order to have a comparison baseline, the proposed system, TempNet, focuses on the startle motion patterns (similarly to \cite{mcintosh2020movement}) observed in sablefish (\textit{Anoplopoma fimbria}), see the example startle event frames of Figure~\ref{fig:1-example}. Prevalence of sablefish startle behaviours can be useful to biological studies of population stress levels. These events are also challenging for computer vision techniques due to their short duration, often in less than 0.5s, and the existence of visually similar behaviours such as regular direction changes or predation events. Startle events can be identified as sudden rapid changes in the speed and trajectory of sablefish and are abundantly observed in video data collected by Ocean Networks Canada\footnote{www.oceannetworks.ca/}. \par

=We propose an efficient novel deep learning-based event detection system, TempNet, which operates on 4-second video clips to detect organism behaviour. Our detector utilizes temporal attention modules based on previous channel attention modules, residual convolutional blocks, wavelet down-sampling, and a custom architecture to improve performance over previous methods while achieving faster than real-time efficiency \cite{woo2018cbam, McIntosh_2021_CRV}.
TempNet not only outperforms the baseline method proposed in \cite{mcintosh2020movement}, but since it does not rely on species-specific visual features, it can be applied to any other behavioural event for which sufficient annotations have been compiled. To the best of our knowledge, our system debuts generic, deep learning-based video detection of fish behaviours (as opposed to the sablefish-specific approach of \cite{mcintosh2020movement}, the single image fish-state detection method of  \cite{HU2021115051}, or the detection-tracking approach of \cite{WANG2022106512}). Figure~{\ref{fig:1-overall}} illustrates the computational pipelines of the proposed detector. \par 

The remainder of this article is structured as follows. In Section \ref{sec:previous_works} we discuss works that are relevant to the proposed detection system. The systems we created for the detection of startle behaviour in sablefish are detailed in Section~\ref{sec:proposed_approach}. We perform comparisons with a state-of-the-art event detection method \cite{yu2018remotenet} and a benchmark hybrid method \cite{mcintosh2020movement} to highlight the performance of TempNet in Section \ref{sec:results}. Additionally, we argue on Section \ref{sec:discussion} that each of the proposed approaches presents particular advantages and disadvantages, thus recommending their use in particular scenarios. \par


\section{Related Works}
\label{sec:previous_works}

Due to the growing domains where large-scale video collections are taking place, such as surveillance and marine environmental monitoring, there has been significant research interest in efficient methods of classifying video and video clips \cite{McIntosh_2021_CRV}, \cite{yu2018remotenet}. Methods like those presented by Bhardwaj \etal~\cite{Bhardwaj_2019_CVPR} focus on reducing the number of frames necessary for video classification. This is done by using a sizeable inefficient large classifier as the teacher to a small student frame-sparse classifier \cite{Bhardwaj_2019_CVPR}. Using this approach, the authors could approach the teacher's performance while reducing the number of Floating Point Operations (FLOPs) by 90\%. Similarly, Tran \etal~\cite{Tran_2019_ICCV} increased model efficiency using channel-separated convolutional neural networks. Channel-separated or group convolutions sit in-between fully interconnected convolutional layers and sparsely connected depth-wise convolutional layers, reducing the number of total convolutional kernels and FLOPs \cite{Tran_2019_ICCV}. The authors found these group convolutions were effective for both 3D and 2D configurations at reducing model FLOPs while maintaining performance \cite{Tran_2019_ICCV}. The proposed methods generated models which were 2-3 times more efficient than comparable models, while retaining state-of-the-art performance \cite{Tran_2019_ICCV}. \par   


\remotenet, a video-based event detector proposed by Yu \etal~\cite{yu2018remotenet}, served as an important inspiration for our proposed system. This lightweight framework fully leverages the spatio-temporal relationships between objects in temporally close frames. Instead of separately considering motion and appearance (as in \cite{saha2016deep,xu2015learning}), \remotenet~ uses 3D CNNs (or ``spatio-temporal attention modules'') to jointly model these video characteristics under the same trainable network. A frame differencing process is executed at the beginning of the pipeline so that the network is encouraged to focus exclusively in the motion-triggered regions of the input. The authors claim that this simple architecture allows for a significant reduction in processing times while preserving the accuracy of its state-of-the-art counterparts \cite{yu2018remotenet}. Among the advantages of \remotenet~is the fact that it is context-agnostic, and it does not require precise, bounding box-based object annotation in videos; instead, this detector uses entire clips as representatives of a certain event class.  

McIntosh \etal~\cite{mcintosh2020movement} offered a hybrid approach that combined object detection, the measurement of four handcrafted features (representing a fish's trajectory direction and speed, bounding box aspect ratio, and rate of local intensity change). Then, a recurrent neural network uses these features to classify 4-second clips as containing sablefish startle events or not. The authors utilize a training data set containing sablefish startle events that have been annotated on both a clip- and specimen-level. Although efficient, the method in \cite{mcintosh2020movement} has two significant disadvantages: a high computational cost (mainly due to the calculation of the handcrafted features) and a strong species-dependency. Our proposed approach tackles these two drawbacks by offering a novel, end-to-end, lightweight deep neural network that outperforms the baseline \cite{mcintosh2020movement} on the dataset introduced while also offering a high level of scalability. \par

\section{Proposed approach} 
\label{sec:proposed_approach}

\begin{figure*}
\begin{center}
\includegraphics[width=1\linewidth]{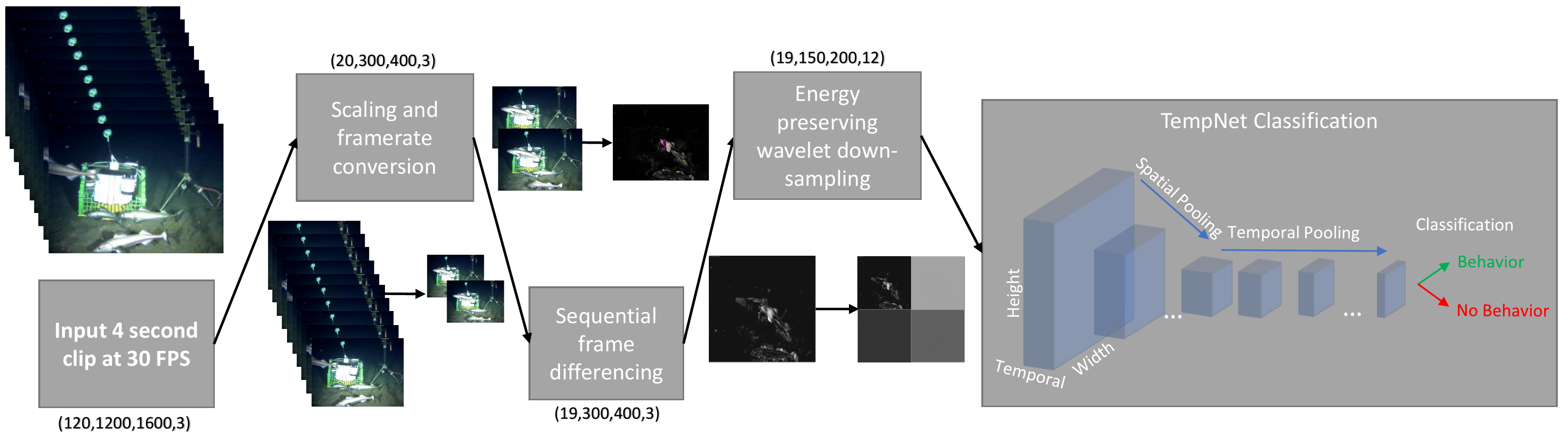}
\end{center}
   \caption{  Proposed approach, TempNet pipeline, for the end-to-end event detection of startle motion in sablefish. The framerate of input clips is first reduced to 5 and an initial scaling takes place. After a frame differencing stage, we employ a Discrete Wavelet Transform-based method~\cite{McIntosh_2021_CRV} to further downsample each frame without sacrificing any of their contents. Tensor output sizes at each stage of the proposed pipeline are presented in the format \textit{(frames, height, width, channels).}
   }
\label{fig:1-overall}
\end{figure*}

Our event detection-based method for biological analysis, TempNet, uses CNNs to determine if a given 4-second clip possesses instances of startle motion of sablefish. TempNet adapts the channel attention proposed in \cite{woo2018cbam} to \textit{temporal} attention to focus the model's architecture on the domain of event detection. TempNet is structured with two distinct encoding phases: one for spatial encoding and the second for temporal encoding. TempNet also uses a bridge region where both spatial and temporal pooling is performed to increase model accuracy, similar to \remotenet \cite{yu2018remotenet}. This sequential encoding forces the model to be deeper with more pooling layers than one-shot encoding. Residual blocks are used to mitigate gradient decay from the increased model depth. From \remotenet~\cite{yu2018remotenet}~we use frame differencing to remove static backgrounds in the clips \cite{yu2018remotenet}. Moreover, we use a Discrete Wavelet Transform (DWT) in a down-sample pre-processing step. A summary of methods can be seen in Figure~\ref{fig:1-overall}.
We use the sablefish clips dataset introduced in \cite{mcintosh2020movement} to train our system and compare the performance of our method with that of other works in the same problem domain. 

\remotenet~was proposed as a detector for specific objects in home surveillance videos \cite{yu2018remotenet}. However, its direct usage in the more challenging task of identifying biological behaviours from the sablefish dataset \cite{mcintosh2020movement} yielded a poor performance (see Table~\ref{tab:4-remoteResults}). We propose a novel architecture:
\begin{enumerate}
    \item A Spatio-Temporal Pooling Bridge where both temporal and spatial pooling are used together, decreasing model depth and improving efficiency, reference Sec~\ref{subsec:sequential}.
    \item Residual blocks to improve training of our relatively deep network (two sequential encoders), see Sec~\ref{subsec:residual_model} \cite{he2016deep}
    \item A temporal attention module during the spatial encoding stage of our model for adaptive feature refinement of the temporal axis before the temporal encoding stage, as in Sec~\ref{subsec:attention}. 
    \item An energy-preserving step~\cite{McIntosh_2021_CRV} for image down-sampling which increases the number of input channels but preserves all of its original high frequency information (otherwise partially lost), as detailed in~\ref{sec:wavelet}. 
\end{enumerate}

\begin{figure}[h]
\centering
\includegraphics[width=0.7\linewidth]{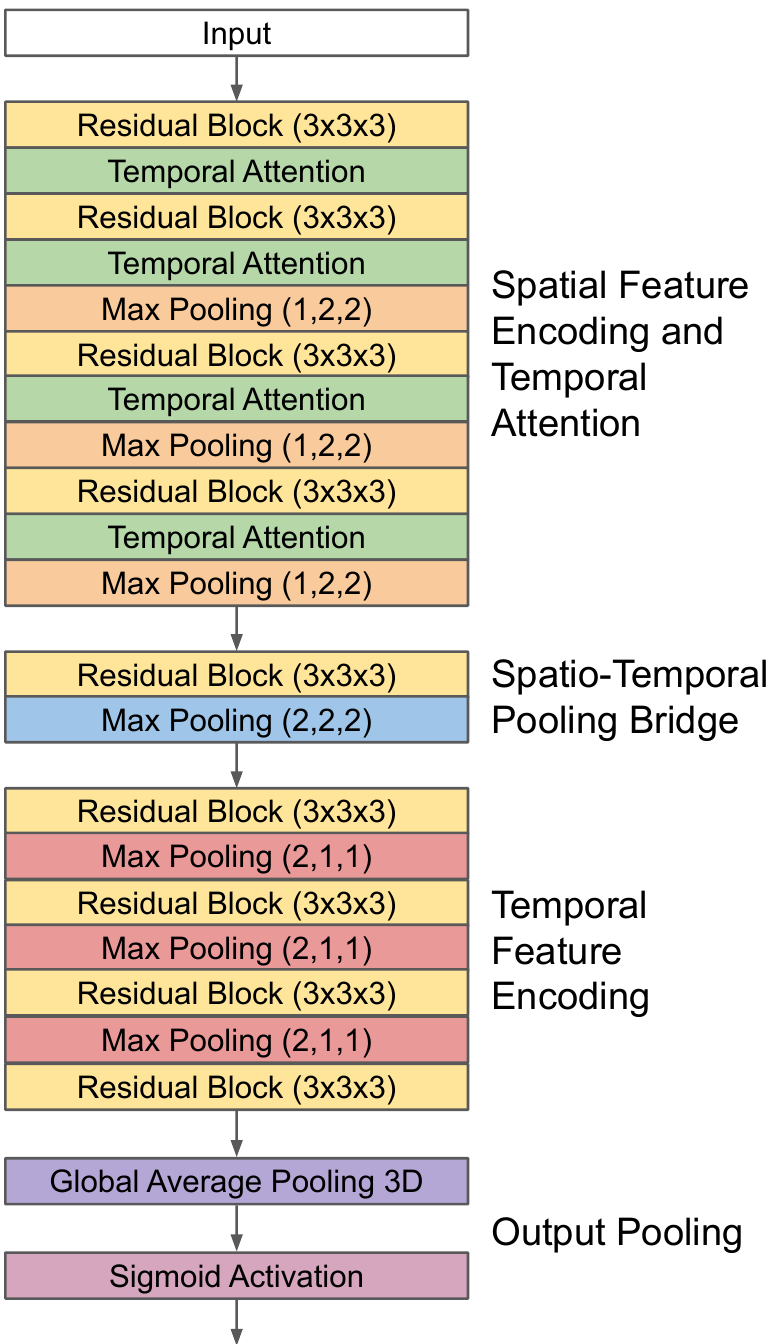}
\caption{Full TempNet architecture. Note the separate stages of the encoder, focusing on spatial encoding and feature extraction then focusing on temporal encoding and feature extraction. }
\label{fig:TempNet}
\end{figure}

Distinctions also include removing \remotenet's use of attention output supervised by the coarse output of a YOLOv3 detector for the objects of interest~\cite{yu2018remotenet}. This change makes our training pipeline simpler to implement and faster to operate, not requiring the supervision of a second detection pipeline. This was found to have little effect on final performance. We maintain some key influences of \remotenet, specifically frame differencing to remove backgrounds in the videos.
The proposed novel considerations are expanded in the following subsections. Our experiments show that the proposed system offers an increased generalization ability, ultimately boosting its accuracy in the sablefish startle motion detection task compared to the original \remotenet~architecture and the comparison baseline~\cite{mcintosh2020movement}. The model architecture is shown in Fig.~\ref{fig:TempNet}.



\subsection{Sequential Encoding and Spatio-Temporal Pooling Bridge}
\label{subsec:sequential}

Similarly to \remotenet,~we use a two-staged encoder \cite{yu2018remotenet}. The first stage of our network performs encoding of the spatial axis, and the second stage encodes the temporal axis. Each stage consists of alternating residual convolutional blocks and max-pooling in the spatial or temporal axes. This sequential architecture is well suited for event detection as spatial features (e.g. the existence of species of interest) should be extracted before temporal features (e.g. how those key objects are moving). The initial encoding of only spatial information promotes the detection of features related to objects to the problem domain (e.g. fish that perform the target behaviour) to be determined by the network. After these spatial features are determined, the temporal stage can extract features in the video to identify behaviour. This sequential encoding of spatial and temporal features increases the necessary depth of the network significantly. The Spatio-Temporal Pooling Bridge region performs a single pooling operation in all spatial and temporal axes, reducing the network's overall depth. This bridge is performed only at the end of the spatial encoding, not to degrade model feature extraction by early temporal pooling. 

\subsection{Residual Blocks} 
\label{subsec:residual_model}

We use Residual Blocks to mitigate convergence issues from the increased depth of the proposed network \cite{he2016deep}. Residual connections allow for additional data paths in networks for information to travel, reducing the gradient degradation noted in deeper networks and improving inter-channel dependencies. This is especially helpful in TempNet due to the two-fold encoding stages, which force the network to be twice as deep. Residual connections also help reduce model size while maintaining high accuracy \cite{he2016deep}. 

In TempNet, we use a standard configuration of Residual Blocks; each block consists of two 3x3x3 convolutions in series with an identity skip connection summed together \cite{he2016deep}. Each convolutional layer in our network has 16 channels to decrease model size.

\subsection{Temporal Attention} 
\label{subsec:attention}

For TempNet, we employ an adapted channel attention model from the convolutional block attention model presented by Woo \etal~\cite{woo2018cbam}. Attention modules have been shown to be simple and computationally efficient ways to improve model representations and, therefore, performance. The adapted formulation from Woo et. al gives a temporal attention map of $M_T(F) \in \mathbb{R} ^{T\times 1\times 1 \times 1}$ generated by the multi-layer perception for an input feature of $F \in \mathbb{R} ^{T\times H\times W \times C}$. The temporal attention coefficients contained in $M_T(F)$ are multiplied element wise to $F$ to garner the attention-scaled temporal frames, in the format $  F' = M_T(F)\circ F  $. These attention coefficients are in the range [0,1] due to the sigmoid activation function used \cite{woo2018cbam}.

\begin{figure}[h]
\centering
\includegraphics[width=0.8\linewidth]{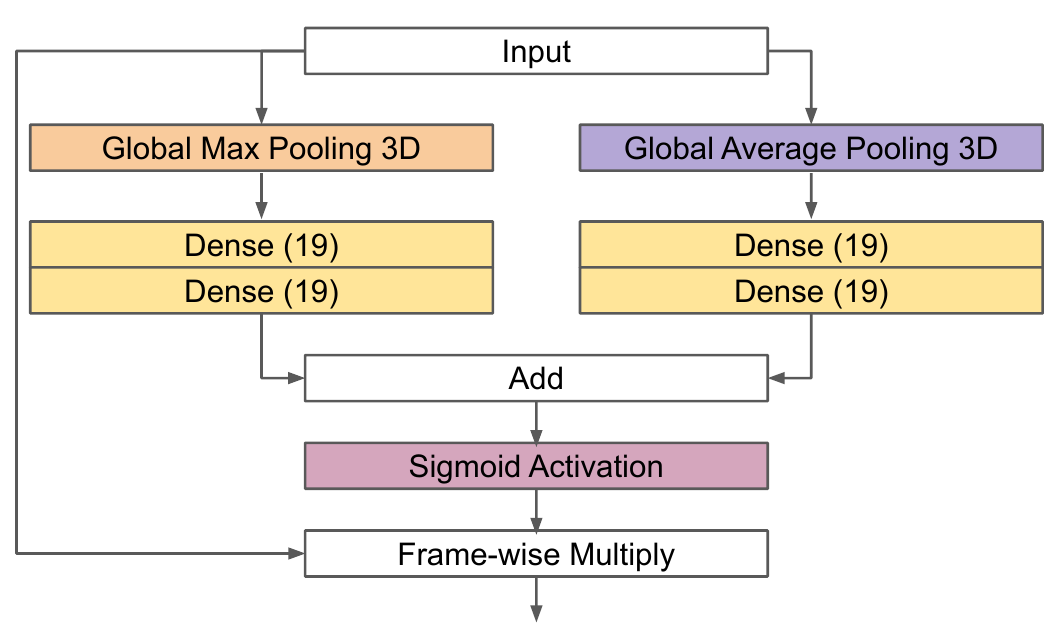}
\caption{Temporal Attention Module. The weights are shared between both MLPs (pairs of dense layers in the figure) as in the channel attention module of CBAM \cite{woo2018cbam}.}
\label{fig:attention}
\end{figure}

We propose the adaptation of this method as a temporal attention module rather than a channel attention module for the domain of event detection. The full adapted attention module can be seen in Fig.~\ref{fig:attention}. We use the temporal attention module to detect specific discrete events within video clips to focus the network. This attention to specific temporal regions is important as the target behaviours are short-lived, often less than 4 frames. Temporal attention is performed on the output in only the spatial encoding steps to ensure that attention mechanisms are performed before any temporal pooling. This ensures that the temporal attention modules can focus the network towards frames which should be kept through the max pooling.

\subsection{Energy Preserving Wavelet Down-Sampling} 
\label{sec:wavelet}
During our preprocessing steps (see Fig.~\ref{fig:1-overall}), after the frame rate conversion (from 29 down to 5 fps) and frame differencing, the system must resize each frame to fit the spatial dimensions required by the network (\textit{i.e.,} 150 $\times$ 200 pixels). 
This common step is often performed using anti-aliasing strategies (\textit{e.g.}~Gaussian filtering followed by sub-sampling) that eliminate high-frequency components of the input. 
In order to achieve the necessary resizing without neglecting high-frequency components, DWTs have been recently employed~\cite{McIntosh_2021_CRV} before the first layers of CNNs. 
Studies show that this strategy can significantly increase the performance of CNN-based tasks where higher resolution images are available~\cite{McIntosh_2021_CRV}. 

\begin{figure}[h]
\centering
\includegraphics[width=0.9\linewidth]{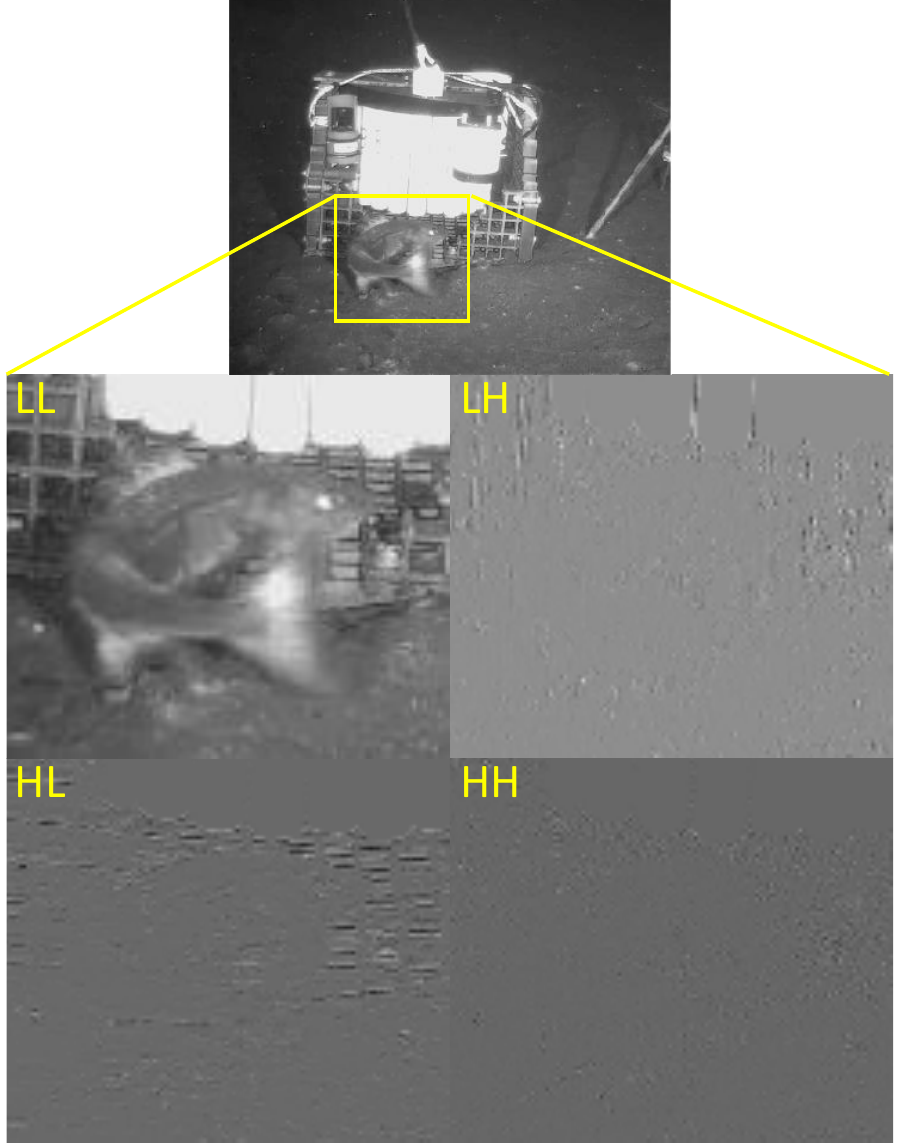}
\caption{Sample output of the DWT based down-sampling on a grey-scale input image. The frequency channel labels are labeled with the vertical component then the vertical component. Note the vertical or horizontal high frequency (e.g. edge) information persevered in the \textit{HL} and \textit{LH} channels respectively.}
\label{fig:wavelet}
\end{figure}

These wavelets create multiple representations of the input that carry all its information and coherently represent the information through the separation of frequency components. These additional input channels increase the number of input channels by a factor of 4 but only increase the number of model parameters and FLOPS by less than 1\% \cite{McIntosh_2021_CRV}, see Table~\ref{tab:3-remoteResults}. This small cost provides the network with the information of an image with 4 times higher spatial resolution than otherwise. Examples of these high-frequency channels can be seen in Figure~\ref{fig:wavelet}. \par

\section{Experimental Results} 
\label{sec:results}

\subsection{Dataset}
\label{subsec:dataset}
McIntosh \etal~\cite{mcintosh2020movement} originally proposed the dataset used in this paper for sablefish startle event detection in short video clips. Video clips in this dataset are acquired at 5fps and contain 20 frames, i.e. 4 seconds of underwater footage. This dataset consists of a total of 892 clips with an even number of positive and negative samples. 100 clips are reserved for testing, 150 for validation. 
The resolution of each sample is 1600$\times$1200 pixels. The data was annotated manually for startle events in each sablefish track (``track-wise'') and the existence of a startle event in the overall clip (``clip-wise''); our paper focuses only on the clip-wise annotations, as we aim to determine if a clip contains startle events or not. 

\subsection{Existing Benchmarks} 

Previous benchmarks presented in \cite{mcintosh2020movement} together with the sablefish dataset detailed the results of a base-level  \remotenet~\cite{yu2018remotenet}~implementation and a novel hybrid system that used deep learning and handcrafted features to identify fish trajectories and the overall presence of startle events. The baseline \remotenet~performance was notably low, whereas the hybrid method proposed in~\cite{mcintosh2020movement} outperformed it by more than 5\% of accuracy, as detailed in Table~\ref{tab:4-remoteResults}. \par
The hybrid method of \cite{mcintosh2020movement} is computationally expensive. For each frame, it requires 1) a complete object detection process (custom YOLOv3~\cite{redmon2018yolov3}), 2) a set of four handcrafted features to be calculated per track, 3) the use of a Long Short Term Memory (LSTM) classifier. Moreover, given that it calculates handcrafted features specific to startle events in sablefish, it is not scalable to other species or problem domains without novel handcrafted features for novel applications.  

\begin{table}[h]
\begin{center}
\begin{tabular}{p{0.35 \linewidth}|P{0.15\linewidth}P{0.14\linewidth}P{0.2\linewidth}}
\hline
Network Structure & Parameters (Thousands) & FLOPs (Millions) & 4s Clip Inference Time  (Milliseconds)*\\
\hline\hline
\mbox{Hybrid LSTM Method \cite{mcintosh2020movement}}  & 61,889    &    128.6  &  3028 \\ \hline
ReMotENet \cite{yu2018remotenet} & 64	&	0.96 & 19.5\\ \hline
TempNet + No Attention & 109  &   1.64 & 38  \\ \hline
TempNet & 110  &   1.64 & 39  \\ \hline
TempNet + Wavelet \\ Down-sampling & 110  &   1.64 & 42  \\ \hline

\end{tabular}
\end{center}
\caption{Efficiency of each of the layouts proposed for event detection in videos. Highest speeds were observed in \remotenet, but all methods excluding the previous Hybrid LSTM method perform an order of magnitude above real-time. (*)Inference time evaluations were performed on a GTX 1080ti.} 
\label{tab:3-remoteResults}
\end{table}

\subsection{Performance of TempNet}

The proposed network improved, when compared to both the hybrid approach of \cite{mcintosh2020movement} and \remotenet~\cite{yu2018remotenet}, in the testing accuracy, precision, and Binary Cross Entropy Error (BCE). All evaluation metrics are generated on the test set of the dataset presented in \cite{mcintosh2020movement}. This subsection explores the influence associated with each of the three novel aspects of our proposed system following an ablation study-like approach.

The base TempNet + No Attention using residual blocks and Spatio-Temporal Pooling Bridge shows significant gains over all previous benchmarks.
The proposed model makes significant strides over the previous Hybrid LSTM Method, increasing Accuracy and Precision by 14.9\% and 5.6\%, see Table~\ref{tab:4-remoteResults}~  \cite{mcintosh2020movement}.
The relative BCE test loss has decreased by 17.5\% over the original \remotenet. Accuracy also improved over \remotenet~ by a significant 16\% absolute improvement without the supervision of a second YOLO-V3 Sablefish detector.   
TempNet + No Attention contains architectural changes designed to mitigate degradation of the network gradients due to the depth of the network in \remotenet. This shows that improving the model gradients with residual connections and decreasing the overall model depth significantly improves model performance. 

The improvement obtained from adding the attention to TempNet is more modest than that of the architectural changes. The addition of Temporal Attention did not lead to significant differences in Accuracy or Precision. However, the addition of Temporal Attention did improve on test BCE by 9.6\%. Indicating the model is still generalizing better despite this not being easily observable on the relatively small test set of this dataset.

\begin{table}[h]
\begin{center}
\begin{tabular}{p{0.25\linewidth}|P{0.10\linewidth}P{0.11\linewidth}P{0.045\linewidth}P{0.05\linewidth}P{0.05\linewidth}P{0.05\linewidth}}
\hline
Network Structure & Accuracy & Precision & BCE & False Neg. & False Pos. & F1\\
\hline\hline

Hybrid LSTM \\Method\cite{mcintosh2020movement} & 0.67  & 0.71 &    1.0  &  21 & 12 & 0.70\\ \hline

ReMotENEt \cite{yu2018remotenet}  & 0.61  & 0.64	 &	 0.63 &  25 & 14 & 0.65\\ \hline
TempNet + \\ No Attention & 0.77    &  0.75 & 0.52 &  9 & 14 & 0.76\\\hline
TempNet & 0.77    &  0.73 & 0.47 &  \bf{8} & 15 & 0.75\\\hline
TempNet+Wavelet \\ Down-sampling & \bf{0.80}    &  \bf{0.81} & \bf{0.46} &  11 & \bf{9} & \bf{0.80}\\
\hline
\end{tabular}
\end{center}
\caption{Performance of each proposed method of event detection and benchmark. Best results are presented in bold.}
\label{tab:4-remoteResults}
\end{table}

Finally, the addition of the Wavelet Down-sample pre-processing improved all metrics. Accuracy was improved over the base TempNet by 3.9\%, while BCE decreased by 2\%. This increase is in line with previous results reported by the work of \cite{McIntosh_2021_CRV}. The best overall results of 80\% accuracy and 0.46 BCE were found using TempNet with Temporal Attention and Wavelet Down-sampling pre-processing. 

Notably, while increasing performance, the full TempNet + Wavelet Down-sampling method is less efficient than \remotenet. This is primarily caused by the architecture change, with an increase of 0.58 million FLOPs, or a 60\% increase. This garners significant model accuracy improvements. However, the entire TempNet pipeline still operates at 476 frames per second, making it applicable to processing large data backlogs. Furthermore, the addition of attention only increased inference time by 1ms. Similarly, the addition of the wavelet Down-sampling increased the overall inference time by 3ms with the CPU-based implementation of the DWT used for this testing, see Table~\ref{tab:3-remoteResults}.

\section{Discussion} 
\label{sec:discussion}

\textbf{Model performance.} Our experimental results show a clear improvement in Accuracy, Precision, and BCE when comparing the full TempNet with ReMotENet~\cite{yu2018remotenet} or the hybrid method of~\cite{mcintosh2020movement}. However, we note that the hybrid method metrics are measured after converting that method's track-wise predictions to clip-wise predictions (as described in the original paper \cite{mcintosh2020movement}), which may degrade its relative performance.

The proposed method produces 25\% fewer false positives (with respect to the hybrid method). This can prove especially useful in analyzing unique and rare biological behaviours in large video datasets where high false-positive rates lead to overwhelming false positives relative to true positives. Our model also yields a reduction of 47.6\% in the number of false negatives relative to the other approaches. We stress that the rate of false negatives is particularly important in applications where specific behaviours or events are sparsely present in a dataset; that is the case with unique and biologically relevant sablefish behaviours such as startles and predation events.

\textbf{Model efficiency.} A lower processing time was a core advancement of the original ReMotENet method, allowing for it to operate beyond real-time on modern hardware \cite{yu2018remotenet}. Differently from the end-to-end CNN event detection-based methods, the hybrid method of McIntosh~\etal~\cite{mcintosh2020movement} required approximately 3s to process each 4-second clip, rendering the processing of years-worth of videos with this method unfeasible. The proposed method is less efficient than the baseline comparison \remotenet. This is primarily due to the architectural changes increasing the number of FLOPs by 60\%. While both methods are designed for large datasets and greater than real-time performance due to the complexity of the behaviour detection, these architectural changes are justified by the gains in performance, see Table~\ref{tab:3-remoteResults}. Each component of the proposed method adds some processing time to each clip but comes with an almost linear improvement in key performance metrics relative to the baseline \remotenet. 

Given their end-to-end CNN structures, the architectures proposed in this work also boast greater than real-time processing capabilities. More specifically, our slowest configurations of TempNet (with Wavelet Down Sampling) can process 4-second videos in 42 ms. This efficiency enables quick processing of large existing marine video datasets and the real-time processing of multiple key data streams for collecting ecologically relevant information~\cite{rountree2019towards}. 

\textbf{Model scalability}. The hybrid method of McIntosh~\etal~\cite{mcintosh2020movement} is based on several environmental assumptions (\eg the trajectory patterns of startle behaviour) and handcrafted features specific to sablefish. Moreover, this method is based on an initial object detection task that looks for instances of the ``sablefish'' class. Similarly, \remotenet~ requires the course annotations of an object detector for the objects of interest for the event \cite{yu2018remotenet}. These species- and event-specific characteristics prevent the simple application of the hybrid and \remotenet~ methods with other species or events.  


Conversely, the proposed method is fully scalable as it employs end-to-end CNNs that do not require any manual feature engineering or supervising detectors, relying solely on appropriate data and annotations. Moreover, a simple modification to the architecture of the proposed system would allow it to identify potentially multiple different behaviours (again, given that enough annotated data is provided for each behaviour).

\section{Conclusion} 
\label{sec:conclusion}

This work offers an autonomous interpretation modality for underwater videos that aims to perform tasks of a higher semantic value in identifying complex biological behaviours. The proposed method is capable of greater than real-time performance, which is a requirement with the growing marine underwater video datasets collected at multiple research institutions. Although we use the startle of sablefish as an initial proof of concept, our proposed event detection-based method is easily scalable to other species and behaviours conditional to sufficient data.\par

The proposed TempNet pipeline debuts four debuts contributions to event detection architectures. These modifications strengthen the detector's capacity to establishing temporal relationships between the 3D CNN-based features, as illustrated by the increased performance reported in our experimental analysis. We evaluate our system in a dataset of startle clips~\cite{mcintosh2020movement} and show that it comfortably outperforms \remotenet~\cite{yu2018remotenet} and the hybrid system of McIntosh~\etal~\cite{mcintosh2020movement} in terms of accuracy and binary cross-entropy error. \par

\bibliographystyle{IEEEtran}
\bibliography{egbib}


\end{document}